\documentclass[11pt,a4paper]{article}
\usepackage[hyperref]{acl2021}
\usepackage{times}
\usepackage{latexsym}

\aclfinalcopy
\def\aclpaperid{1090} 

\usepackage[utf8]{inputenc} 
\usepackage{microtype}
\usepackage{graphicx}
\usepackage{subfigure}
\usepackage{booktabs} 
\usepackage{comment}
\usepackage{hyperref}
\usepackage{mathtools}
\usepackage{color}
\usepackage{graphicx}
\usepackage{CJKutf8}
\usepackage{url}
\usepackage{multicol,multirow}
\usepackage{arydshln}
\usepackage{makecell}
\usepackage{array}
\usepackage{bm}
\usepackage{cancel}
\usepackage{enumitem}
\usepackage{parskip}
\usepackage[T1]{fontenc}    
\usepackage{booktabs}       
\usepackage{amsmath, amsthm}
\usepackage{amssymb}
\usepackage{amsfonts}
\usepackage[ruled]{algorithm2e}

\usepackage{enumitem}

\usepackage{microtype}
\definecolor{ao}{rgb}{0.0, 0.5, 0.0}
\definecolor{asparagus}{rgb}{0.53, 0.66, 0.42}
\definecolor{amber}{rgb}{1.0, 0.49, 0.0}
\definecolor{alizarin}{rgb}{0.82, 0.1, 0.26}
\definecolor{applegreen}{rgb}{0.55, 0.71, 0.0}
\definecolor{amethyst}{rgb}{0.6, 0.4, 0.8}
\definecolor{auburn}{rgb}{0.43, 0.21, 0.1}
\def\aclpaperid{1090} 

\title{BertGCN: Transductive Text Classification \\ by Combining GCN and BERT}

\author{Yuxiao Lin$^\spadesuit$, Yuxian Meng$^\clubsuit$, Xiaofei Sun$^\clubsuit$\\
{\bf Qinghong Han$^\clubsuit$,  Kun Kuang$^\spadesuit$,
Jiwei Li$^{\spadesuit\clubsuit}$ and Fei Wu$^\spadesuit$}\\
$^\spadesuit$College of Computer Science and Technology, Zhejiang University\\ 
$^\clubsuit$ ShannonAI \\
\{yuxiaolinling, kunkuang, jiwei\_li, wufei\}@zju.edu.cn\\
\{yuxian\_meng, xiaofei\_sun, qinghong\_han\}@shannonai.com
}

\begin{document}
\maketitle

\begin{abstract}
In this work, we propose BertGCN, a model that combines large scale pretraining and transductive learning for text classification. BertGCN constructs a  heterogeneous graph over the dataset and represents documents as nodes using BERT representations. By jointly training the BERT and GCN modules within BertGCN, the proposed model is able to leverage the advantages of both worlds: large-scale pretraining which takes the advantage of the massive amount of raw data and transductive learning which jointly learns representations for both training data and unlabeled test data by propagating label influence through graph convolution. Experiments show that BertGCN achieves SOTA performances on a wide range of text classification datasets.\footnote{Code available at \url{https://github.com/ZeroRin/BertGCN}.}
\footnote{Accepted by Findings of ACL2021.}
\end{abstract}

\section{Introduction}
Text classification is a core task in natural language processing (NLP) and has been used in many real-world applications such as spam detection \citep{wang2010don} and opinion mining \citep{bakshi2016opinion}.
Transductive learning \citep{Vapnik1998} is a particular method for text classification which makes use of both labeled and unlabeled  examples in the training process.
Graph neural networks (GNNs) serve as an effective approach for transductive learning \citep{yao2019graph,liu2020tensor}. 
In these works, a graph is constructed to model the relationship between documents. Nodes in the graph represent text units such as words and documents, while edges are constructed based on the semantic similarity between nodes. GNNs are then applied to the graph to perform node classification. 
The merits of GNNs and transductive learning are as follows:
(1) the decision for an instance (both training and test) does not depend merely on itself, but also its neighbors. This makes the model more immune to data outliers; 
(2) at the training time, since the model propagates influence from supervised labels across both training and test instances through graph edges,  unlabeled data also contributes to the process of representation learning, and consequently higher performances.   


Large-scale pretraining has recently demonstrated their effectiveness on a variety of NLP tasks \citep{devlin2018bert, yinhan2019roberta}. 
Trained on large-scale unlabeled corpora in an unsupervised manner, 
large-scale pretrained models are able to learn  implicit but rich text semantics in language at scale. 
Intuitively, large-scale pretrained models have potentials to benefit transductive learning.
 However, existing models for transductive text classification \citep{yao2019graph,liu2020tensor} did not take large-scale pretraining into consideration, and its effectiveness  still remains unclear.


In this work, we propose BertGCN, a model that combines the  advantages of both large-scale pretraining and transductive learning for text classification. BertGCN constructs a heterogeneous graph for the  corpus with node being word or document, and  node embeddings are initialized with pretrained BERT representations, and uses graph convolutional networks (GCN) for classification.
By jointly training the BERT and GCN modules,  the  proposed  model  is  able  to  leverage the advantages of both worlds: large-scale pretraining which takes the advantage of the massive amount of raw data and transductive learning which jointly learns representations for both training data and unlabeled test data by propagating label influence through graph edges.
The proposed BertGCN model successfully combines the powers of large-scale pretraining and graph networks, and achieves new state-of-the-art performances on a wide range of   text classification datasets.

\section{Related Work}
Graph neural networks (GNNs) are connectionist models that capture dependencies and relations between graph nodes via message passing through edges that connect nodes \citep{scarselli2008graph,hamilton2017inductive,xu2018powerful}. GNNs are practically categorized into \citep{wu2020comprehensive}: graph convolutional networks \citep{kipf2016semi,wu2019simplifying}, graph attention networks \citep{velivckovic2017graph,zhang2018gaan}, graph auto-encoder \citep{cao2016deep,kipf2016variational}, graph generative networks \citep{de2018molgan,li2018learning} and graph spatial-temporal networks \citep{li2017diffusion,yu2017spatio}. 
GNNs serve as powerful tools to utilize the relationship between different objects, and have been applied to various domains such as traffic prediction \citep{yu2018spatio,zhang2018gaan} and recommendation \citep{zhang2020comprehensive,monti2017geometric}.
In the context of NLP, GNNs have achieved remarkable successes across a wide range of end tasks such as relation extraction \citep{zhang2018graph}, semantic role labeling \citep{marcheggiani-titov-2017-encoding}, data-to-text generation \citep{marcheggiani-perez-beltrachini-2018-deep}, machine translation \citep{bastings-etal-2017-graph} and question answering \citep{song2018exploring,de2018question}.

The prevalence of neural networks has motivated a diverse array of works on developing neural models for text classification. Different neural model architectures \citep{kim2014convolutional,zhou2015c,radford2018improving,chai2020description} have demonstrated their effectiveness against traditional statistical feature based methods \citep{wallach2006topic}. 
Other works leverage label embeddings and jointly train them along with input texts \citep{wang2018joint,pappas2019gile}.
More recently, the success achieved by large-scale pretraining models has spurred great interests in adapting the large-scale pretraining framework \citep{devlin2018bert} into text classification \citep{reimers2019sentence}, leading to remarkable progressive on few-shot \citep{mukherjee2020uncertaintyaware} and zero-shot \citep{ye-etal-2020-zero} learning. 

Our work is inspired by the work of using graph neural networks for text classification \citep{yao2019graph,huang2019text,zhang2020text}. But different from these works, we focus on combining large-scale pretrained models and GNNs, and show that GNNs can significantly benefit from large-scale pretraining. 
Existing works that combine BERT and GNNs uses graph to model relationships between tokens within a single document sample \citep{lu2020vgcn,he2020enhancing}, which fall into the category of inductive learning. Different from these works, we use graph to model relationships between different samples from the whole corpus to utilize the similarity between labeled and unlabeled documents, and uses GNNs to learn their relationships.

\section{Method}

\subsection{BertGCN}

In the proposed BertGCN model, 
we initialize  representations for document nodes in a text graph 
using
 a BERT-style model (e.g., BERT, RoBERTa).
 These representations 
   are used as inputs to GCN. 
 Document  representations will then be iteratively updated based on the graph structures  using GCN, the outputs of which are treated as final representations for document nodes, 
 and are sent to the softmax classifier for predictions. 
In this way,
 we are able to  leverage the complementary strengths of pretrained models and graph models.


Specifically, we construct a heterogeneous graph containing both word nodes and document nodes following TextGCN \citep{yao2019graph}. We define word-document edges and word-word edges based on the term frequency-inverse document frequency (TF-IDF) and positive point-wise mutual information (PPMI), respectively. The weight of an edge between two nodes $i$ and $j$ is defined as:
\begin{equation}
\small
    \begin{split}
    A_{i,j}= \left \{
        \begin{array}{ll}
            \operatorname{PPMI}(i, j), & i, j\text{ are words and } i \neq j \\
            \operatorname{TF-IDF}(i, j),     & i\text{ is document, }j\text{ is word}\\
            1, & i = j\\
            0, & \text{otherwise}
        \end{array}
    \right.
    \end{split}
    \label{adj_mat}
\end{equation}
In TextGCN, an identity matrix $X = I_{n_\text{doc}+n_\text{word}}$ is used as initial node features, where $n_\text{doc}$ is the number of document nodes, $n_\text{word}$ is the number of word nodes (including both training and test). 
In BertGCN, we use a BERT-style model  to obtain the  document embeddings, and treat them as input representations for document nodes. 
Document node embeddings are denoted by $X_\text{doc}\in\mathbb{R}^{n_\text{doc}\times d}$, where $d$ is the embedding dimensionality.
Overall, the initial node feature matrix is given by:
\begin{equation}
    X= \left (
        \begin{array}{c}
            X_\text{doc}\\
            0
        \end{array}
    \right )_{(n_\text{doc}+n_\text{word})\times d}
\end{equation}
We feed $X$ into a GCN model \citep{kipf2016semi} which iteratively propagates messages across training and test examples. Specifically, the output feature matrix of the $i$-th GCN layer $L^{(i)}$ is computed as
\begin{equation}
    L^{(i)} = \rho(\tilde A L^{(i-1)} W^{(i)})
\end{equation}
where $\rho$ is an activation function, $\tilde A$ is the normalized adjacency matrix and $W^{(i)}\in \mathbb{R}^{d_{i-1}\times d_i}$ is a weight matrix of the layer. $L^{(0)}=X$ is the input feature matrix of the model.
Outputs of GCN are treated as final representations for documents, which is then fed to the softmax layer for classification:
\begin{equation}
Z_\text{GCN} = \operatorname{softmax}(g(X, A))
\label{gcn}
\end{equation}
where $g$ represents the GCN model. 
We use the cross entropy loss over labeled document nodes to jointly optimize parameters for BERT and GCN.
 
 \subsection{Interpolating BERT and GCN Predictions}
Practically, we find that optimizing BertGCN with a auxiliary classifier that directly operates on BERT embeddings leads to faster convergence and better performances. 
Specifically, we construct an auxiliary classifier by directly feeding document embeddings (denoted by $X$) to a dense layer with softmax activation:
\begin{equation}
  Z_\text{BERT} = \operatorname{softmax}(WX)
\end{equation}
The final training objective is the linear interpolation of the prediction from BertGCN and the prediction from BERT, which is given by:
\begin{equation}
    Z = \lambda Z_\text{GCN} + (1-\lambda) Z_\text{BERT}
\end{equation} 
where $\lambda$ controls the tradeoff between the two objectives. 
$\lambda=1$ means we use the full BertGCN model, and $\lambda=0$ means we only use the BERT module. When $\lambda\in (0,1)$, we are able to balance the predictions from both models, and the BertGCN model can be better optimized.

The explanation for better performances achieved by the interpolation is as follows: 
The $Z_\text{BERT}$ directly operates on the input of GCN, making sure that inputs to GCN are regulated and optimized towards the  objective.
This helps the multi-layer GCN model to overcome intrinsic drawbacks such as gradient vanishing or over-smoothing \citep{li2018deeper}, and thus leads to better performances. 

 \subsection{Optimization using Memory Bank}
 

The original GCN model uses the full-batch gradient descent method for training, which is intractable for the proposed BertGCN model, since the full-batch method can not be applied to BERT due to the memory limitation.
Inspired by techniques in contrastive learning which decouples the dictionary size from the mini-batch size \citep{wu2018unsupervised,he2020momentum}, we introduce a memory bank that stores all document embeddings to decouple the training batch size from the total number of nodes in the graph.
 
Specifically, during training, we maintain a memory bank $M$ that tracks input features for all document nodes. At the beginning of each epoch, we first compute all document embeddings using the {\it current} BERT module and store them in $M$. During each iteration, we sample a mini batch from both labeled and unlabeled document nodes with the index set $B = \{b_0, b_1...b_n\}$, where $n$ is the mini-batch size. We then compute their document embeddings $M_B$ also using the {\it current} BERT module and update the corresponding memories in $M$.\footnote{Note that the BERT module used to compute $M_B$ is the one finished training in the last iteration, which is different from the the BERT module used to compute the initial $M$.}
Next, we use the updated $M$ as input to derive the GCN output and compute the loss for the current mini batch. For back-propagation, $M$ is considered as constant except the records in $B$.

With the memory bank, we are able to efficiently train the BertGCN model including the BERT module.
However, during training, the embeddings in the memory bank are computed using the BERT module at different steps in an epoch and are thus inconsistent. To overcome this issue, we set a small learning rate for the BERT module to improve consistency of the stored embeddings. With low learning rate the training takes more time. In order to speed up training, we fine-tune a BERT model on the target dataset before training begins, and use it to {\bf initialize} the BERT parameters  in  BertGCN.

\section{Experiments}
\subsection{Experiment Setups}
We run experiments on five widely-used  text classification benchmarks: 20 Newsgroups (20NG)\footnote{\url{http://qwone.com/˜jason/20Newsgroups/}}, R8 and R52\footnote{\url{https://www.cs.umb.edu/~smimarog/textmining/datasets/}}, Ohsumed\footnote{\url{http://disi.unitn.it/moschitti/corpora.htm}} and Movie Review (MR)\footnote{\url{http://www.cs.cornell.edu/people/pabo/movie-review-data/}}. 

We compare BertGCN to current state-of-the-art pretrained and GCN models: TextGCN \citep{yao2019graph}, SGC \citep{wu2019simplifying}, BERT \citep{devlin2018bert} and RoBERTa \citep{yinhan2019roberta}. Details for datasets and baseline are left in the supplementary material.

We follow protocols in TextGCN to preprocess data. 
For BERT and RoBERTa, we use the output feature of the [CLS] token as the document embedding, followed by a feedforward layer to derive the final prediction. We use BERT$_\text{base}$ and a two-layer GCN to implement BertGCN. We initialize the learning rate to 1e-3 for the GCN module and 1e-5 for the fine-tuned BERT module. We also implement our model with RoBERTa and GAT \citep{velivckovic2017graph}. GAT variants are trained over the same graph as GCN variants, but learn edge weights through attention mechanism instead of using pre-defined weight matrix.

\subsection{Main Results}

\begin{table}
  \small
  \centering
  \scalebox{0.85}{
  \begin{tabular}{lccccc}\toprule
    {\bf Model} & {\bf 20NG} & {\bf R8} & {\bf R52} & {\bf Ohsumed} & {\bf MR}\\
    \midrule 
    {\it TextGCN}  &  86.3& 97.1& 93.6& 68.4 &76.7\\
    {\it SGC} & 88.5 &97.2 &94.0 & 68.5 & 75.9\\
    \cdashline{1-6}
    {\it BERT} & 85.3 & 97.8 & 96.4 & 70.5 & 85.7\\
    {\it RoBERTa} & 83.8 & 97.8 & 96.2 & 70.7 & 89.4\\
    \cdashline{1-6}
    {\it BertGCN} & 89.3  & 98.1 & \bf{96.6} & \bf{72.8} & 86.0 \\
    {\it RoBERTaGCN} & {\bf 89.5} & {\bf 98.2} & 96.1 & {\bf 72.8} & {\bf 89.7}\\
    {\it BertGAT} & 87.4 & 97.8 & 96.5 &  71.2  & 86.5 \\
    {\it RoBERTaGAT} & 86.5 & 98.0 & 96.1 &  71.2  & 89.2 \\
    \bottomrule
  \end{tabular}
  }
  \caption{Results for different models on transductive text classification datasets. We run all models 10 times and report the mean test accuracy.}
  \label{tab:main}
\end{table}

Table \ref{tab:main} presents the test accuracy of each model. We can see that BertGCN and RoBERTaGCN perform the best across all datasets. Only using BERT and RoBERTa generally performs better than  GCN variants except 20NG, which is due to the great merits brought by large-scale pretraining. 
Compared with BERT and RoBERTa, the performance boost from BertGCN and RoBERTaGCN is significant on the 20NG and Ohsumed datasets. This is because the average length in 20NG and Ohsumed is much longer than that in other datasets:
the graph is constructed using word-document statistics, which means that long texts may produce more document connections transited via an intermediate word node, and this potentially benefits message passing through the graph, leading to better performances when combined with GCN.
This may also explain why GCN models perform better than BERT models on 20NG.
For datasets with shorter documents such as R52 and MR, the power of graph structure is limited, and thus the performance boost is smaller relative to 20NG. 
BertGAT and RoBERTaGAT can also benefit from the graph structure, but their performance are not as good as GCN variants due to the lack of edge weight information.

\subsection{The Effect of $\lambda$}
\begin{figure}
    \centering
        \includegraphics[width=0.43\textwidth]{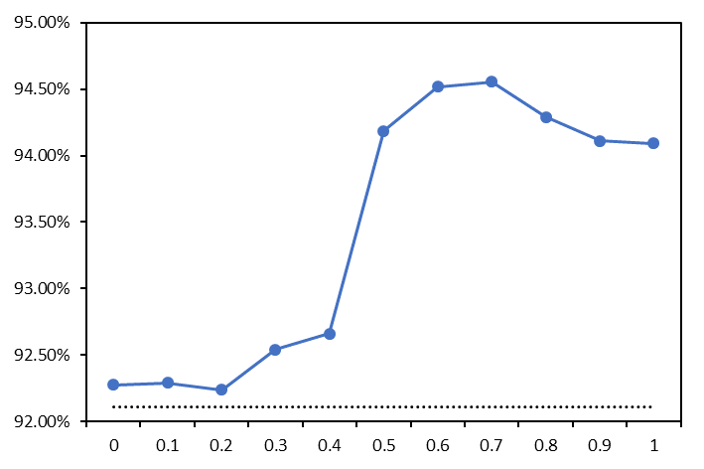}
    \caption{Accuracy of RoBERTaGCN when varying $\lambda$ on 20NG development set. The dotted line indicates the corresponding RoBERTa baseline.\footnotemark}
    \label{fig:lambda}
\end{figure}
\footnotetext{The original training/test split of 20NG is based on post date, but the development set is randomly sampled from the original training set. The accuracy on test set is thus much lower than that on development set.}

$\lambda$ controls the trade-off between training BertGCN and BERT. The optimal value of $\lambda$ can be different for different tasks. Fig.\ref{fig:lambda} shows the accuracy of RoBERTaGCN with different $\lambda$. On 20NG, the accuracy is consistently higher with larger $\lambda$ value. This can be explained by the high performance of graph-based methods on 20NG. 
The model reaches its best when $\lambda=0.7$, performing slightly better than only using the GCN prediction ($\lambda=1$). 

\subsection{The Effect of Strategies in Joint Training }
\begin{table}
  \small
  \centering
  \scalebox{0.85}{
  \begin{tabular}{lcccc}\toprule
    {\bf Strategy} & {\it w/ both} & {\it w/o finetune} & {\it w/o small lr.} & {\it w/o both}\\
    {\bf Accuracy} & 94.7 & 93.8 & 10.3\footnotemark & 10.3\textsuperscript{\ref{converge}}\\
    \bottomrule
  \end{tabular}}
  \caption{Accuracy on 20NG development set for different strategies. ``finetune'' means we use the finetuned RoBERTa as initialization, and ``small lr.'' means we use a smaller learning rate for the RoBERTa module.}
  \label{tab:ab_init_lr}
\end{table}
\footnotetext{Experiments without a small lr. failed to converge.\label{converge}}

To overcome inconsistency of embeddings in the memory bank, we set a smaller learning rate for the BERT module and use a finetuned BERT model for initialization. We evaluate the effect of the two strategies. Table \ref{tab:ab_init_lr} shows the  results of RoBERTaGCN on 20NG with and without these strategies. 
With the same learning rate for RoBERTa and GCN, the model cannot be trained due to inconsistency in the memory bank, regardless of whether the fine-tuned RoBERTa is used. Models can be successfully trained when we set a smaller learning rate for the RoBERTa module, and additional using finetuned RoBERTa leads to the best performance.

\section{Conclusion and Future Work}
In this work, we propose BertGCN, which takes the best advantages from both large-scale pretraining models and transductive learning for text classification. We efficiently train BertGCN by using a memory bank that stores all document embeddings and updates part of them with respect to the sampled mini batch. 
The framework of BertGCN can be built on top of any document encoder and any graph model. 
Experiments demonstrate the power of the proposed BertGCN model. However, in this work, we only use document statistics to build the graph, which might be sub-optimal compared to models that are able to automatically construct edges between nodes. We leave this in future work.

\section*{Acknowledgement}
This work is supported by National Key R\&D Program of China (2020AAA0105200) and Beijing Academy of Artificial Intelligence (BAAI).

\bibliography{emnlp2020}
\bibliographystyle{acl_natbib}

\appendix
\section{Dataset Details}
\begin{itemize}
    \item The 20NG dataset\footnote{\url{http://qwone.com/˜jason/20Newsgroups/}} contains 18,846 newsgroups posts from 20 different topics. We use the bydate version which splits the dataset to 11,314 train samples and 7,532 test samples based on the posting date.
    \item R8 and R52\footnote{\url{https://www.cs.umb.edu/~smimarog/textmining/datasets/}} are two subsets of the Reuters dataset with respectively 8 and 52 categories. R8 has 5,485 training and 2,189 test documents. R52 has 6,532 training and 2,568 test documents.
    \item The OHSUMED test collection\footnote{\url{http://disi.unitn.it/moschitti/corpora.htm}} is a set of references from MEDLINE, the online medical information database. Following previous works,  we use 7,400 documents belonging to one of the 23 disease categories to form a classification dataset, with 3,357 documents for training and 4,043 for test.
    \item MR \citep{pang-lee-2005-seeing}\footnote{\url{http://www.cs.cornell.edu/people/pabo/movie-review-data/}} is a movie review dataset for binary sentiment classification. The corpus has 10,662 reviews. We use the train/test split in \citet{tang2015pte}
\end{itemize}

\section{Baselines}
\begin{itemize}[noitemsep]
  \item TextGCN \citep{yao2019graph}: TextGCN is a model that operates graph convolution over a word-document heterogeneous graph. Node features are initialized using an identity matrix.
  \item SGC \citep{wu2019simplifying}: Simple Graph Convolution is a variant of GCN that reduces the complexity of GCN by removing non-linearities and collapsing weight matrices between consecutive layers.
  \item BERT \citep{devlin2018bert}: BERT is a large-scale pretrained NLP model.
  \item RoBERTa \citep{yinhan2019roberta}: a robustly optimized BERT model that improves upon BERT with different pretraining methods.
\end{itemize}

\end{document}